\documentclass[a4paper, 10pt, conference]{ieeeconf}      %

\IEEEoverridecommandlockouts                              %

\usepackage[caption=false,font=footnotesize]{subfig}
\usepackage{dblfloatfix}
\usepackage{graphics} %
\usepackage{graphicx}
\usepackage{epsfig} %
\usepackage{amsmath} %
\usepackage{amssymb}  %
\usepackage{amsfonts}
\usepackage{color}
\usepackage{subfig}
\usepackage{adjustbox}
\usepackage{multirow}
\graphicspath{{Figures/}}
\usepackage{import}
\usepackage{color,soul}
\title{\LARGE \bf
Soft Continuum Actuator Tip Position and Contact Force Prediction, Using Electrical Impedance Tomography and Recurrent Neural Networks
}

\author{Amirhosein Alian$^{1}$, George Mylonas$^{1}$, and James Avery$^{1}$ %
\thanks{This work was partially supported by a collaboration with the Multi-Scale Medical Robotics Centre, The Chinese University of Hong Kong. James Avery is an Imperial College Research Fellow. For the purpose of open access, the authors have applied a Creative Commons Attribution (CC BY) license to any Accepted Manuscript version arising.
}%
\thanks{$^{1}$The Hamlyn Centre, Imperial College London, London W2 1NY, UK. \textit{Corresponding author: Amirhosein Alian} {\tt\small a.alian21@imperial.ac.uk}}
}

\begin{document}

\maketitle
\thispagestyle{empty}
\pagestyle{empty}

\begin{abstract}
Enabling dexterous manipulation and safe human-robot interaction, soft robots are widely used in numerous surgical applications. One of the complications associated with using soft robots in surgical applications is reconstructing their shape and the external force exerted on them. Several sensor-based and model-based approaches have been proposed to address the issue. In this paper, a shape sensing technique based on Electrical Impedance Tomography (EIT) is proposed. The performance of this sensing technique in predicting the tip position and contact force of a soft bending actuator is highlighted by conducting a series of empirical tests. The predictions were performed based on a data-driven approach using a Long Short-Term Memory (LSTM) recurrent neural network. The tip position predictions indicate the importance of using EIT data along with pressure inputs. Changing the number of EIT channels, we evaluated the effect of the number of EIT inputs on the accuracy of the predictions. The least RMSE values for the tip position are 3.6 and 4.6 mm in Y and Z coordinates, respectively, which are 7.36\% and 6.07\% of the actuator's total range of motion. Contact force predictions were conducted in three different bending angles and by varying the number of EIT channels. The results of the predictions illustrated that increasing the number of channels contributes to higher accuracy of the force estimation. The mean errors of using 8 channels are 7.69\%, 2.13\%, and 2.96\% of the total force range in three different bending angles.
\end{abstract}

\section{Introduction}
\label{sect:1}

Soft robots are widely used in many fields and applications such as minimally invasive surgery (MIS) \cite{Cianchetti2014}, terrain navigation \cite{Li2022}, rehabilitation \cite{Thalman2018}, and industrial grippers \cite{Tawk2019}. Soft robots provide compliance, facilitate dexterous manipulation, and ensure safe human-robot interaction. Despite considerable benefits associated with the use of soft robotics, certain challenges remain to be addressed. One of these challenges is state estimation during actuation and interaction with the environment. The high compliance of soft robots enables them to deform nonlinearly under external and internal loads, thus making the prediction of their states, such as tip position and backbone shape, demanding. In surgical applications, the presence of anatomical barriers and unstructured environments make state estimation even more challenging. Accurate state estimation contributes to lower pain and discomfort in patients \cite{Szura2012} and reduces postoperative complications and adverse events during operations \cite{Eickhoff2010, So2021}. Estimating exerted forces by soft robots can contribute equally to patients’ comfort. A large number of studies have investigated force prediction from shape sensing modules of soft robots \cite{Gao2020, Xu2017, Qiao2021}. 
Apart from intraoperative imaging, model-based and sensor-based approaches have been investigated thus far to address the state estimation challenges in the field of soft surgical robots. %

In sensor-based approaches, Fiber Bragg Gratings (FBGs), ElectroMagnetic (EM) trackers, and stretchable sensors are primarily used for state estimation. FBGs are MRI compatible and biocompatible optical sensors that provide state information of the robot with high sampling rates \cite{So2021, Sefati2021}, measuring the wavelength shifts of the emitted light.
Although FBGs intrinsically have limited range of motion, recent studies have shown using Nitinol wires in combination with FBGs can enhance their deformability \cite{Roodsari2022}. FBGs require a bulky and expensive optical spectrum interrogator to measure the wavelength shift \cite{Park2010}. Additionally, the distance between the interrogator and optical cores adversely influences the accuracy, thus posing limitations on the application of FBGs in surgical devices \cite{Xiao2018}. EM trackers have minimum impact on mechanical characteristics of the overall system, however the  electromagnetic field of operating rooms can severely impact performance. Generally used in flexible robots with multiple serially connected sections \cite{Shi2017}, several studies estimated the shape and tip position of flexible robots through fitting Bezier curves to EM tracker data \cite{Song2015, Song2015Electro}. Stretchable sensors composed of conductive materials are another method used for state estimation of soft robots by measuring resistance \cite{Truby2020}, capacitance \cite{Bilodeau2018}, or inductance \cite{Xing2020} changes induced by the deformations of the device. The limitations of stretchable sensors are hysteresis, requiring additional wiring, and material biocompatibility concerns \cite{Sahu2021}. %

Continuing the study conducted in \cite{Avery2019} where EIT was shown effective in deformation detection, this paper evaluates the tip position and force estimation accuracy of a data-driven approach using EIT data. 
Biocompatibility and MRI compatibility are two potential benefits of this sensing approach in inflatable medical devices. 
Compared to previously mentioned sensing schemes, EIT shape sensing requires minimal wiring, has negligible effect on system physical characteristics, and minimizes manufacturing and sensing integration costs. To employ this sensing technique, a conductive liquid is used to pressurise the soft actuator, which induces a deformation and volume change within the chamber. Using an array of electrodes embedded inside the actuation chamber, electrical current is injected between different regions and the resultant voltages are recorded. The changes in these voltages are attributed to the changes in the volume of the liquid and shape changes of the actuator. Leveraging the measured voltages in combination with a recurrent neural network, we evaluate the performance of EIT sensing in tip position prediction and contact force estimation. Since EIT images provide no information of the sequence of the data, a data-driven approach was adopted to predict the states based on the arrays of voltage changes. Experiments are executed using a single degree of freedom (DOF) soft bending actuator to validate the accuracy of EIT state estimation in a data-driven manner. An OptiTrack camera system and a 6-axis force transducer are used for training the network and as the ground truth for the EIT state estimation assessment. As voltages are recorded through several pairs of electrodes, the effect of number of measurement pairs on the accuracy of state estimation of EIT is evaluated in this study. The proposed sensing scheme can be potentially used in flexible endoscopy and laparoscopy where the shape changes of the end effector can be measured while meeting safety concerns.   %

\begin{figure}
\centering
\includegraphics[width=1\columnwidth]{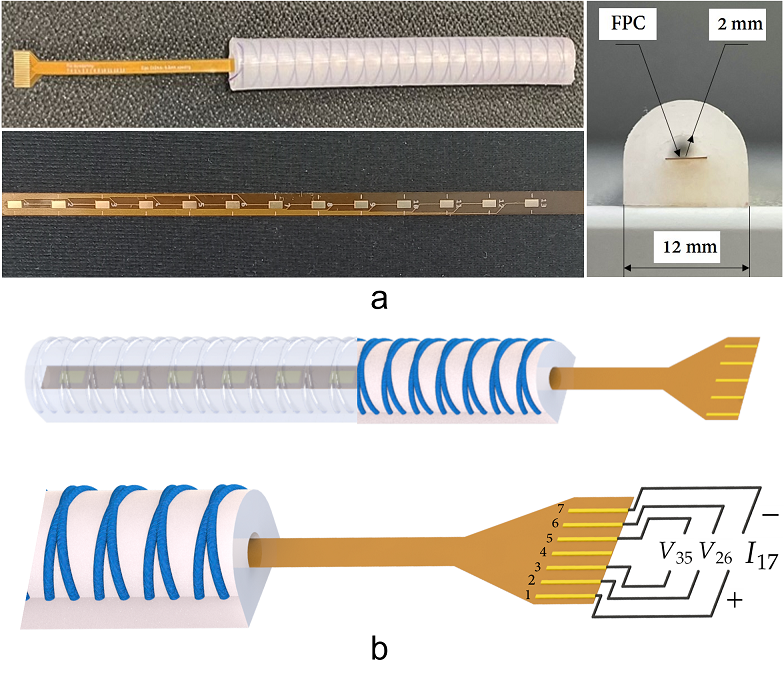}
\caption{a) The soft continuum actuator and the integrated FPC containing 13 electrodes. The FPC was placed on the flat side of the semi-circle cross-section of the actuator. b) Schematic of the integrated actuator with an 8-electrode FPC, and how the impedances are measured. While current $I_{17}$ is injected, $V_{26}$ , $V_{35}$ are recorded simultaneously. Subscripts refer to  electrode number, so $I_{17}$ is the current injected between electrodes 1 and 7.}
\label{fig:1}
\end{figure}

\begin{figure*}
\centering
\includegraphics[width=2\columnwidth]{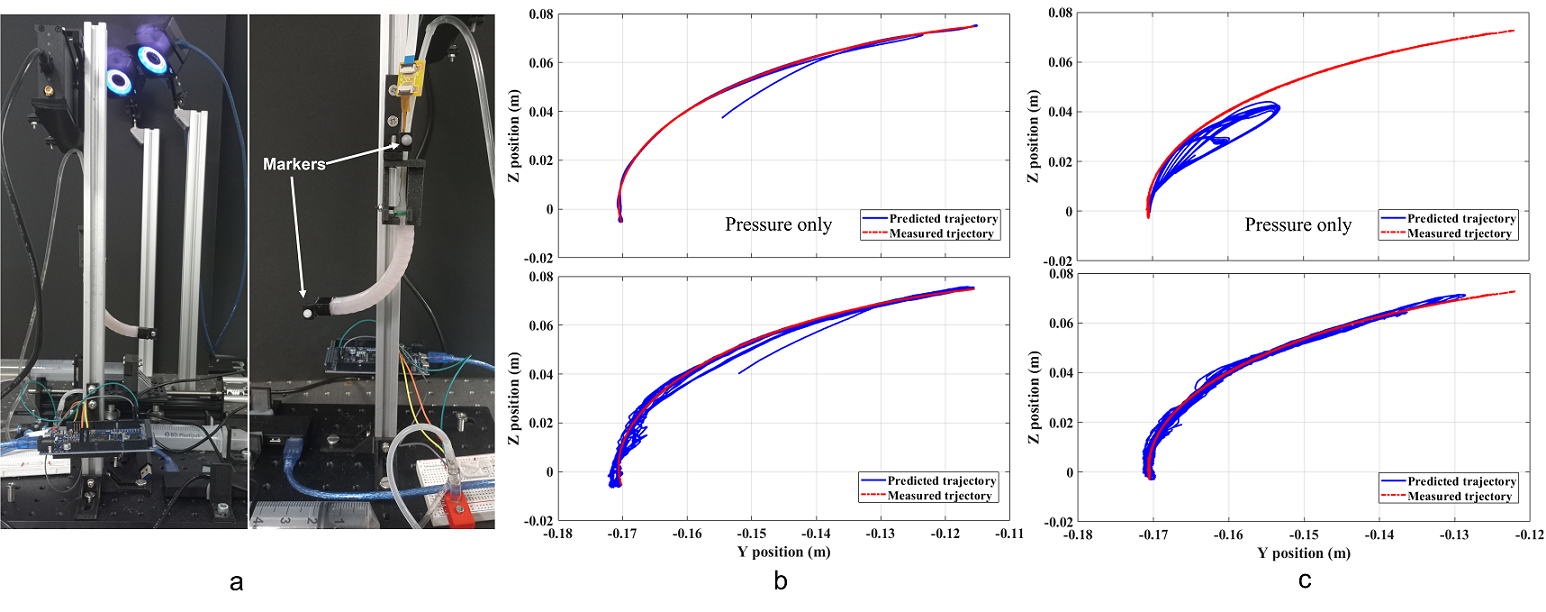}
\caption{a) The setup used for tip position estimation using motion capture system. The effect of using pressure and EIT impedance changes in the accuracy of the tip position estimation upon actuation b) with constant peak amplitude, top: pressure only as input, bottom: pressure and EIT impedance as inputs. c) with random peak amplitude, top: pressure only as input, bottom: pressure and EIT impedance as inputs. }
\label{fig:3}
\end{figure*}
\section{Materials and Methods}
\label{Methods}

\subsection{EIT hardware system }
\label{EIT hardware system }
The primary unit of the EIT system utilized here is Quadra impedance spectroscopy developed by Eliko tech \cite{min2018bioimpedance}. This unit incorporates an alternating current source, voltage acquisition and multiplexer, which enables the recording of impedance data from up to 16 electrodes.
An impedance measurement requires current to be injected through a pair of electrodes, and voltage to be recorded through another pair, a sequence of these measurements is referred to as an EIT protocol.
In this study, current of magnitude 1 mA at 11 kHz frequency was used in all experiments, which is injected sequentially, known as Time Division Multiplexing (TDM), at a rate of 20 Hz for a complete protocol. To estimate the state of the robot, the recorded voltages can be directly exploited to train the network or EIT reconstructions could be used as in \cite{Avery2019}.
\subsection{Fabricating the Soft Actuator}
\label{Fabrication of the soft actuator}
The soft bending actuator is fabricated through casting Ecoflex 00-50 Silicone rubber into a 3-D printed mold following the same procedure discussed in \cite{Polygerinos2015}. The mold is designed such that the actuator has an inflatable chamber with a semi-circle cross-section. Fiber reinforcement and an inextensible polyester layer are added to restrain the radial and axial expansions of the actuator, respectively.
Being able to bend in-plane with a single DOF, the fabricated actuator is 100 mm long with a wall thickness of 4 mm and total diameter of 12 mm. Several of the proposed continuum actuator can be concatenated into a tethered endoscopic device for applications such as gastrointestinal endoscopy.  %
Prior to capping the distal end of the actuator, a Flexible Printed Circuit (FPC) containing the injection and measurement electrodes was integrated into the actuation chamber. The FPC contains 13 electrodes placed linearly with constant spacing of 6.5 mm on a Polyamide flexible film. The flexibility and stiffness of the film has a minimum impact on the mechanical behaviour of the actuator. The electrodes were gold coated to minimize the contact impedance between the conductive fluid and the electrodes, thus reducing the noise in the voltage signals. The fluid used for pressuring the actuator is 0.9\% saline. To integrate the FPC into the actuator, the Polyamide film is placed on the flat side of the semi-circle cross section of the actuator. To fix the position of the FPC and minimize the noise stemming from the displacement of the electrodes, both ends of the FPC are glued using a silicone epoxy. %
The fabricated soft bending robot with the integrated FPC is shown in Fig. \ref{fig:1}.

\subsection{LSTM Network}
\label{Fabrication of the soft actuator}
To predict the tip position and the contact force of the soft actuator an LSTM network is trained using a combination of pressure and EIT voltages. LSTM is a dynamic network capable of learning time series events by keeping a memory of data from past events \cite{Staudemeyer2019}. As soft robots display a time-varying behaviour and the data related to their states is sequential, LSTM is a potential candidate to be used for predicting the tip position and contact force. In the study, the training and predictions are performed by the LSTM used in \cite{Thuruthel2019}. Ultimately, an LSTM network with layer size of 50 and dropout rate of 0.1 is used. The ratio of the unseen data for evaluating the prediction accuracy to the training data is 1 to 4. The training data is normalised by the mean and standard deviation of the training dataset. The network with the same architecture but different training datasets is used for all the predictions in this study. %

\section{Experimental Setup}
\label{Experimental Setup}

The experimental setup used for data collection comprises the soft bending robot, two motion capture systems, a hydraulic pressure sensor, actuation and computation units, and a force/torque transducer. 
Two reflective markers were placed on the rigid structure and at the distal end of the robot for optical tracking of the tip position using OptiTrack motion capture system (see Fig. \ref{fig:3}.a). 
The actuation unit incorporates a hydraulic pump system along with connecting tubes. The hydraulic pump system consists of a leadscrew and stepper motor driven by a uStepper S-lite controller board and a syringe filled with 0.9\% saline. An Arduino Due generated the control signals to pressurize the actuator, and recorded the absolute hydraulic pressure as measured by a MS5803-14BA sensor \cite{Avery2022}. The pressure data was sampled at 20 Hz and recorded in MATLAB through serial communication. 

Two cameras (Prime 13 OptiTrack, NaturalPoint, USA) are employed to track the reflective marker at the tip of the actuator and record its coordinates in 3-D space. OptiTrack data collected at 125 Hz is utilized to train the network and as the ground truth for the tip position predictions. For contact force prediction experiments, an 6-axis force/torque transducer (Nano 17, ATI Automation, USA) is employed. 
The force data sampled at 62.5 Hz is used for training and validation of the LSTM. Impedance changes upon pressurization of the actuator are sampled by Quadra impedance spectroscopy at 20 Hz. Based on the protocol defined with 9 channels, the current is injected between two outlying electrodes (e.g. $I_{17}$ in Fig \ref{fig:1}), and impedance changes are measured through pairs of electrodes in between (e.g. $V_{26}$ and $V_{35}$ in Fig \ref{fig:1}). Since the sampling rate of the capture motion system and the force sensor exceed that of the EIT system, the force and the tip position data are resampled at 20 Hz. %
\begin{figure*}
\centering
\includegraphics[width=2\columnwidth]{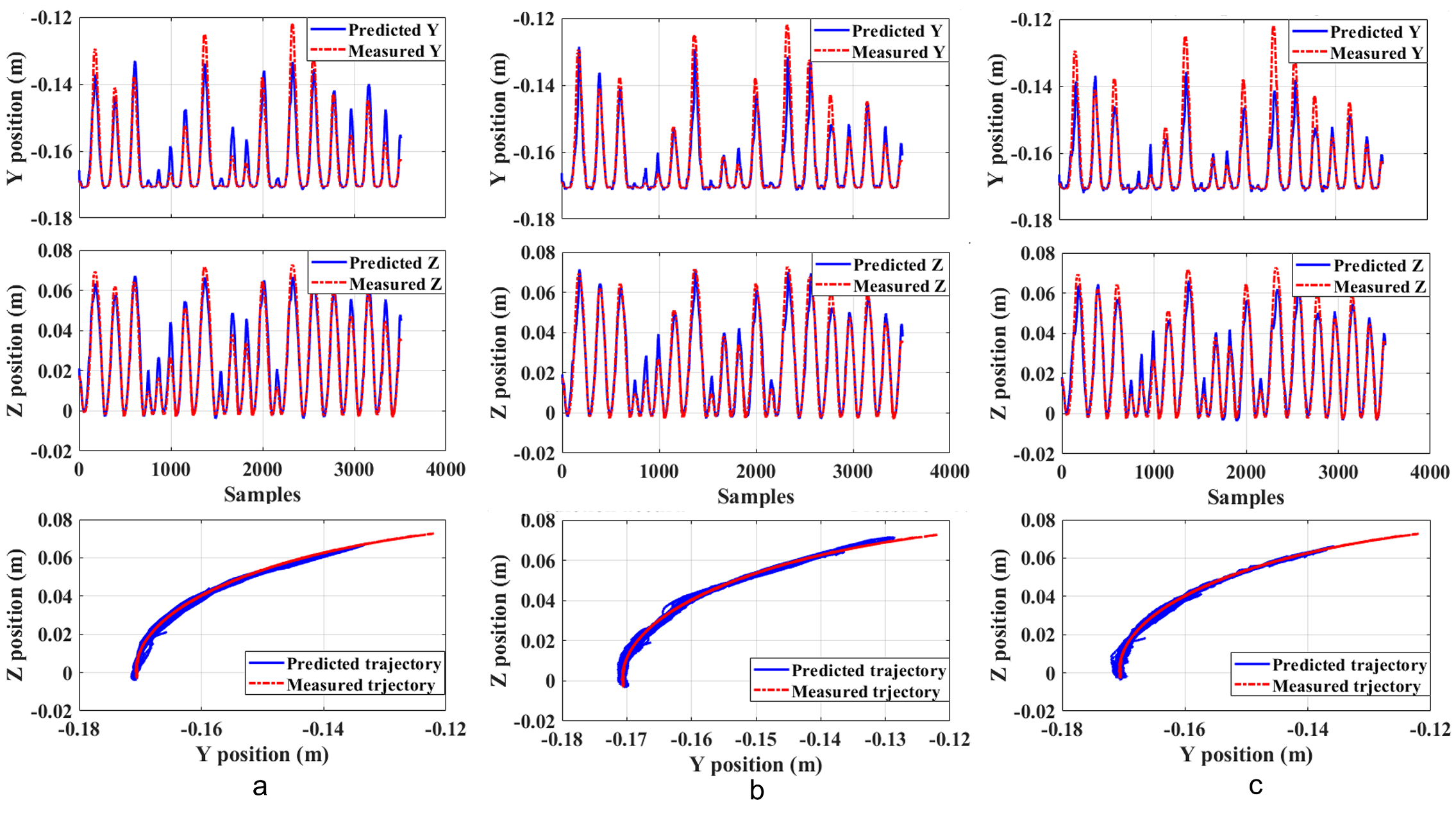}
\caption{The effect of the number of EIT channels used for the tip prediction. a) One channel b) Four channels c) Eight channels. The results indicate that while using 4 channels can lead to enhanced estimation compared to 1 channel, increasing this number to 8 may not have the same effect. The RMSE values for tip estimation with 4 channels are 7.36\% and 6.07\% of the actuator's total range of motion in Y and Z, respectively. }
\label{fig:4}
\end{figure*}
\subsection{Tip Position Estimation}
To evaluate the reliability of the EIT data to predict the tip position of the soft bending robot, two sets of experiments were conducted. First, the actuator was set to bend with a constant period and amplitude of 4000 steps, using a constant velocity of 1500 steps per minute, which corresponded to minimum and maximum pressures of 1.131 and 1.675 bar respectively. The complete dataset was 10 minutes or 58 repetitions. In the second experiment, the amplitudes between 1000 and 5000 steps were chosen randomly while the slope and the velocity remained the same. It was found in initial testing that the extra complexity in this required additional training data, so 20 minutes or 116 actuations were used in this case. In both experiments, the pressure and EIT data are used as the input to the LSTM network, and Y-Z coordinates of the tip position are defined as the output.
The test data used as the ground truth for the predictions incorporates the unseen data from the sampled dataset. To demonstrate the improvements offered by EIT, the tip prediction was performed using pressure data alone, and then in combination with EIT data through four channels. 

To discover the effect of the number of EIT channels used for training the LSTM, the tip position was estimated by training the network using pressure and 3 different subsets of the original EIT dataset containing 1, 4, and 8 EIT channels.

\subsection{Contact Force Estimation}
\begin{figure}
\centering
\includegraphics[width=0.8\columnwidth]{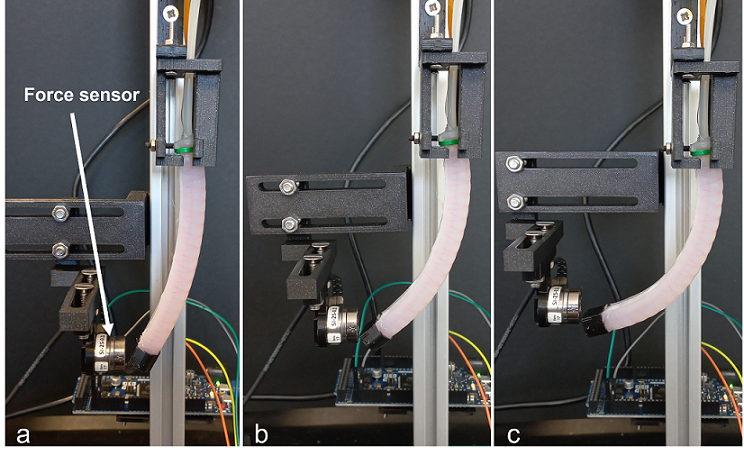}
\caption{Collecting the contact force data for training the network in three different bending angles.}
\label{fig:5}
\end{figure}

To collect the contact force data, the force transducer was placed within the working space of the soft actuator. The position of the sensor in the vertical plane was adjusted using a rigid structure connected to the aluminium stand. The contact force dataset was obtained at three different bending angles of the actuator to evaluate the effect of the shape on the precision of the predictions (see Fig \ref{fig:5}). Before pressurizing the actuator, it was pre-curved such that the tip has slight contact with the sensor. The actuator was pressurized by a triangle wave signal with random peak values.  The data collection at each bending angle lasted 10 minutes followed by resampling the data at 20 Hz EIT sampling rate. The feature variables of the LSTM incorporated the pressure and EIT impedance values of 4 channels while the target variable was set to the force data.
\section{Results}
\label{Results}
\subsection{Tip Position Estimation}
\label{Tip Position Estimation}

Fig. \ref{fig:3} compares the predicted tip position for the networks trained with pressure and with pressure and EIT. In the first experiment with repetitive actuation, using the pressure alone has approximately the same prediction results as using both the pressure and the EIT data (see Fig. \ref{fig:3}.b). This outcome can be intuitively concluded as the data sequence is highly repeatable and thus easier to predict by the network. However, when the pressure signals have random peak amplitudes, the pressure fails to predict the tip position with any accuracy (Fig. \ref{fig:3}.c). %
It is observed that the RMSE of the predictions by the pressure values is smaller than using both the pressure and EIT data in the experiment with non-random pressure. This difference can be explained by the additional noise in the EIT signals, that impacts the accuracy of the predictions. In case of constant peak amplitude, the overall RMSE values in vertical plane are 0.825 and 2.08 mm using pressure only and in combination with EIT data, respectively. However, these values for random peak amplitude are 21.6 and 6.24 mm. %

Regarding the effect of the number of channels, as shown in Fig. \ref{fig:4}, the dataset with 4 EIT channels yielded slightly enhanced predictions than the dataset with 1 channel. However, the prediction accuracy dropped by increasing the number of EIT channels to 8. This can be concluded by calculating the RMSE values. The least RMSE values for the tip position are 3.6 and 4.6 mm in Y and Z coordinates, respectively. These values are 7.36\% and 6.07\% of the actuator's total range of motion in Y and Z, respectively. The overall RMSE values in vertical plane are 5.84, 6.24, and 7.21 mm in case of using 1 channel, 4, and 8 channels, respectively. The results underline the fact that using a higher number of channels does not necessarily enhance the accuracy of the predictions, as the channels are not linearly independent and thus may not contain much additional information, but may introduce further noise. Optimizing the number of channels is important as the higher number can increase the computational cost of the system, and complexity of wiring and manufacture, while it has no contribution to improving the predictions.
\subsection{Contact Force Estimation}
\label{Contact Force Estimation}
\begin{figure}[bp]
\centering
\includegraphics[width=1\columnwidth]{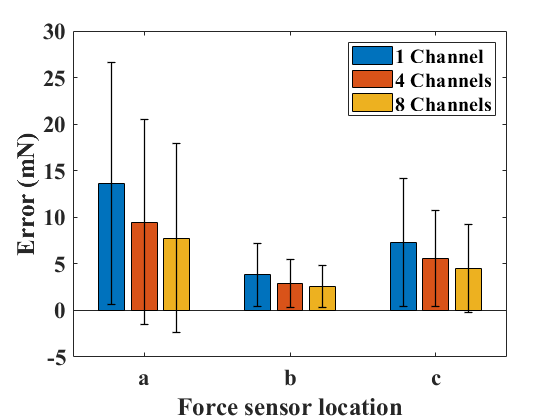}
\caption{Mean and standard deviation for the contact force estimation respective to the number of channels and the location of the force sensor}
\label{fig:7}
\end{figure}

\begin{figure}
\centering
\includegraphics[width=1\columnwidth]{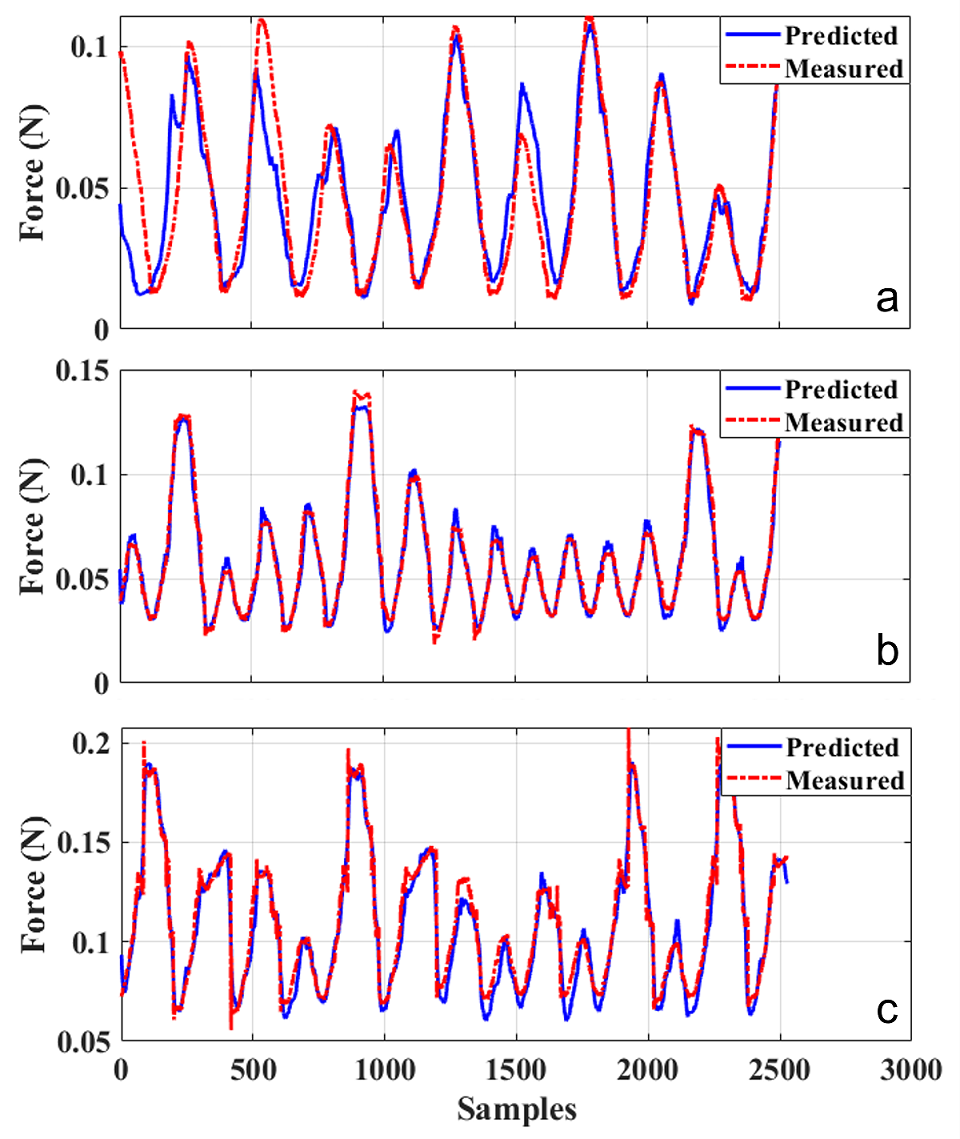}
\caption{The accuracy of the contact force prediction in three different location using 4 channels of EIT.}
\label{fig:6}
\end{figure}

 Using the same LSTM and training to test dataset ratio, the prediction results respective to three bending angles are depicted in Fig. \ref{fig:6}. Further predictions were conducted to evaluate the influence of channel numbers on the results. Therefore, the contact force was predicted using three different number of channels similar to ones used in tip position estimation. These predictions were repeated for all three bending angles. The mean and standard deviation of the absolute error between the ground truth data and predictions are illustrated in Fig. \ref{fig:7}. The force was predicted most accurately in location b with 8 EIT channels, with a mean error of 2.6 mN, or 1.82\% of the maximum force applied. Additionally, regardless of the location of the force sensor, the mean and the standard deviation of the error decline by using higher number of channels for training the network. Higher number of channels provides more detailed information of the actuator’s states during the force exertion. Furthermore, the relatively high prediction error in location (a) can be attributed to the lower Signal to Noise Ratio (SNR) of the impedance data measured in this location. The actuator’s deformation is close to its rest shape in location (a), hence yielding minimal impedance differentials. 

\section{Discussion and Conclusion}
\label{sect:5}
In this paper, the EIT sensing technique with broad applications in the shape sensing and force estimation of medical devices was proposed. Experiments were conducted to evaluate the reliability and effectiveness of EIT in predicting the tip position and contact force of a soft bending actuator. Since the actuator bends in plane, the external force is exerted and estimated in the same plane. However, out of the plane force exertion will be studied in future works with a multi degree of freedom actuator fabricated in \cite{shen2022}.
The results showed using pressure data as the only input of the LSTM leads to noticeably less accurate predictions when the continuum robot is actuated by random values of pressure. The prediction RMSE values for using pressure only were 10.1 and 19.1 mm in Y and Z coordinates, respectively. However, the inclusion of the EIT data in the feature variables resulted in RMSE values of 4 and 4.8 mm in the same coordinates. These values are 8.18\% and 6.33\% of the actuator's range of motion in Y and Z coordinates respectively, which are more promising than some studies such as \cite{Thuruthel2019}. %
Increasing the number of channels used in the tip position predictions can have an undesired effect on the resultant accuracy. Particularly, using only a single channel yielded less erroneous predictions than using 4 and 8 channels in terms of RMSE values. The higher accuracy of a single channel can stem from the fact that one channel is more sensitive to deformations. However, changing the EIT protocol for defining measurement and injection pairs might vary the results. Additionally, higher number of channels is probably required in case of interacting with an obstacle or the exertion of an external load along the length of the actuator. %

The accuracy of the force prediction increases by using a higher number of EIT channels. The mean errors of using 8 channels are 7.8, 2.6, and 4.5 mN, which are 7.69\%, 2.13\%, and 2.96\% of the total force range (improved results compared to \cite{Thuruthel2019}). This value for using 1 channel equals to 13.6, 3.8, and 7.3 mN in location (a), location (b), and location (c), respectively.
It can be observed that the bending angle at which the external force is exerted can have an impact on the accuracy of the force predictions. This can be attributed to SNR value of the impedance signal which is the lowest in location (a). This result is due to the minimal difference between the actuator’s shape in location (a) and its shape in rest position.
In future studies, the proposed EIT system will be used to predict the entire shape of the actuator’s backbone by using additional markers along the actuator and parameterising the curves. Shape estimation will be conducted using different machine learning models and will be compared to LSTM. The prediction of the shape while external forces are being exerted simultaneously and shape will be explored in future works. Additionally, the Quadra impedance spectroscopy will be replaced by a frequency division multiplexing (FDM) method which enhances the sampling rate of the measurements. 
\section*{ACKNOWLEDGMENTS}
This work is partially supported by a research collaboration with the %
Multi-Scale Medical Robotics Centre, The Chinese University of Hong Kong.

\bibliographystyle{IEEEtran}
\bibliography{reference.bib}
 
\end{document}